\documentclass[
hf,
]{ceurart}

\usepackage{graphics}
\usepackage{makecell,tabularx}
\usepackage[vlined, linesnumbered ]{algorithm2e}
\SetKwInput{KwInput}{Input}                % Set the Input
\usepackage{amsmath,amsfonts,amssymb,mathrsfs}
\usepackage{xcolor}
\usepackage{adjustbox}
\usepackage{pdflscape}
\begin{document}

%%
%% Rights management information.
%% CC-BY is default license.
% \copyrightyear{2021}
% \copyrightclause{Copyright for this paper by its authors.
%   Use permitted under Creative Commons License Attribution 4.0
%   International (CC BY 4.0).}

%%
%% This command is for the conference information
\conference{ArXiv preprint}

%%
%% The "title" command
\title{Synthesizing Rolling Bearing Fault Samples in New Conditions: A framework based on a modified CGAN}

\author[1]{Maryam Ahang}[%
email=maryam.ahang@uvic.ca,
]

\author[1,2,3]{Masoud Jalayer}[%
email=masoud.jalayer@polimi.it,
]

\author[1]{Ardeshir Shojaeinasab}[%
]

\author[2]{Oluwaseyi Ogunfowora}[%
]

\author[1]{Todd Charter}[%
]

\author[1,2]{Homayoun Najjaran}[%
email=najjaran@uvic.ca,
]

\address[1]{Department of Electrical and Computer Engineering, University of Victoria, V8P 5C2, Victoria, BC, Canada}
\address[2]{Department of Mechanical Engineering, University of Victoria, Victoria BC, V8P 5C2, Canada}
\address[3]{DIG, Politecnico di Milano, Via Lambruschini 24/b, 20156, Milan, Italy}

%%
%% The abstract is a short summary of the work to be presented in the
%% article.
\maketitle
\begin{abstract}
Bearings are one of the vital components of rotating machines that are prone to unexpected faults. Therefore, bearing fault diagnosis and condition monitoring is essential for reducing operational costs and downtime in numerous industries. In various production conditions, bearings can be operated under a range of loads and speeds, which causes different vibration patterns associated with each fault type. Normal data is ample as systems usually work in desired conditions. On the other hand, fault data is rare, and in many conditions, there is no data recorded for the fault classes. Accessing fault data is crucial for developing data-driven fault diagnosis tools that can improve both the performance and safety of operations. To this end, a novel algorithm based on Conditional Generative Adversarial Networks (CGANs) is introduced. Trained on the normal and fault data on any actual fault conditions, this algorithm generates fault data from normal data of target conditions. The proposed method is validated on a real-world bearing dataset, and fault data are generated for different conditions. Several state-of-the-art classifiers and visualization models are implemented to evaluate the quality of the synthesized data. The results demonstrate the efficacy of the proposed algorithm.
\end{abstract}

%%
%% Keywords. The author(s) should pick words that accurately describe
%% the work being presented. Separate the keywords with commas.

% \newpage

\section*{Nomenclature}
\begin{table}[h]
\centering
\resizebox{0.8\textwidth}{!}{\begin{tabular}{llll}
\toprule
\textbf{Abbreviations} &                                             & VAE               & Variational Autoencoder              \\
AE           & Auto-Encoder                                & WGAN              & Wasserstein GAN                      \\
AI           & Artificial Intelligence                     &                   &                                      \\
CGAN         & Conditional Generative Adversarial Networks & \textbf{Symbols}           &                                      \\
CNN          & Convolutional Neural Networks               & $c_n$             & Working conditions                   \\
CNN          & Convolutional Neural Network                & $f_k$             & Kernel filter                        \\
ConvAE       & Convolutional Auto Encoder                  & $b_k$             & Bias                                 \\
ConvLSTM     & Convolutional LSTM                          & $\sigma$          & Activation function                  \\
CWRU         & Case Western Reserve University             & $\mathscr{P}_{r}$ & Distribution of the raw data         \\
DBN          & Deep Belief Networks                        & $\mathscr{P}_{f}$ & Distribution of the the fake samples \\
EDM          & Electro-Discharge Machining                 & $y$               & Extra information                    \\
GAN          & Generative Adversarial Networks             & $z$               & Random noise vector                  \\
IFD          & Intelligent Fault Diagnosis                 & $G$               & Generative model                     \\
k-NN         & K-Nearest Neighbour                         & $D$               & Discriminator model                  \\
LSTM         & Long Short-Term Memory                      & $f_t$             & Forget gate                          \\
N2FGAN       & Normal to fault GAN                         & $g_t$             & Cell candidate                       \\
RNN          & Recurrent Neural Network                    & $i_t$             & Input gate                           \\
RPM          & Revolutions Per Minute                      & $o_t$             & Output gate                          \\
SVM          & Support Vector Machine                      & $C_t$             & Cell state                           \\
t-SNE        & t-distributed Stochastic Neighbor Embedding & $h_t$             & Hidden state                        \\
\bottomrule
\end{tabular}}
\end{table}
\clearpage
\section{Introduction}
Bearings are a crucial part of rotating machinery and are widely used in many industries; it was recorded that 44\% of machine faults experienced in manufacturing industries are related to bearing failures\cite{m1}. Fault detection is a critical part of system design and maintenance because it helps to improve production efficiency resulting in reduced costs and accidents. Fault detection has gained the interest in academia and industry, and has been a hot topic of research because of its significance \cite{m2}.
Fault detection and diagnosis methods are generally classified into two groups: model-based methods and data-driven methods. In model-based methods, the model's output and the actual system's signals are used to generate several symptoms differentiating normal from faulty machine states; based on these symptoms, faults are determined through classification or inference methods. However, data-driven methods rely on the sensor data collected from the plant and usually use Artificial Intelligence (AI) to learn and classify characteristic fault features from data. AI plays an inevitable role in industry and manufacturing systems \cite{m17}. 
\par Deep learning methods, which are known for their capability to process massive amounts of data and are relatively robust against noise are the best methods for intelligent fault detection \cite{m3}. Convolutional Neural Networks (CNN) \cite{m7,m8}, Stacked Auto Encoders \cite{m4,m5}, and Deep Belief Networks (DBN)\cite{m6}  are among the most studied algorithms that could reach very high accuracy. Industrial machines mainly operate in normal conditions, so there are more normal data than fault data that makes the available data imbalanced. Even though some methods like one-class classification and novelty detection can detect faults in such conditions, identifying the type of faults is not possible \cite{m9}. To solve this problem, Generative algorithms can be employed to generate fault data. Generative algorithms are unsupervised learning paradigms that automatically discover patterns in input data so the model can produce new examples. Variational Autoencoders (VAE) and Generative Adversarial Networks (GAN) are the most famous generative models and have been wildly used for bearing fault detection \cite{m10, m11}, and data augmentation \cite{m19},  and predicting remaining useful life. By using generative algorithms, the problem of lack of samples and patterns in industrial data can be solved \cite{berghout2022systematic}. CGAN is a variation of GAN, that can generate conditional new data \cite{m12}. In \cite{m13}, where datasets are limited and imbalanced, a conditional deep convolutional generative adversarial network is used for machine fault diagnosis. Yin et al. \cite{m14} also applied a data generation method based on Wasserstein generative adversarial network and convolutional neural network for bearing fault detection.  These methods aim to extract the input data's probability distribution and hidden information so that they can be sampled and used to generate new data. Moreover, the distribution of any condition is unique, so the distributions of unknown conditions can be found with the information of known ones and used to generate data for new conditions where fault data is not available.
\par
\begin{figure}
\centering
\includegraphics[width=\textwidth]{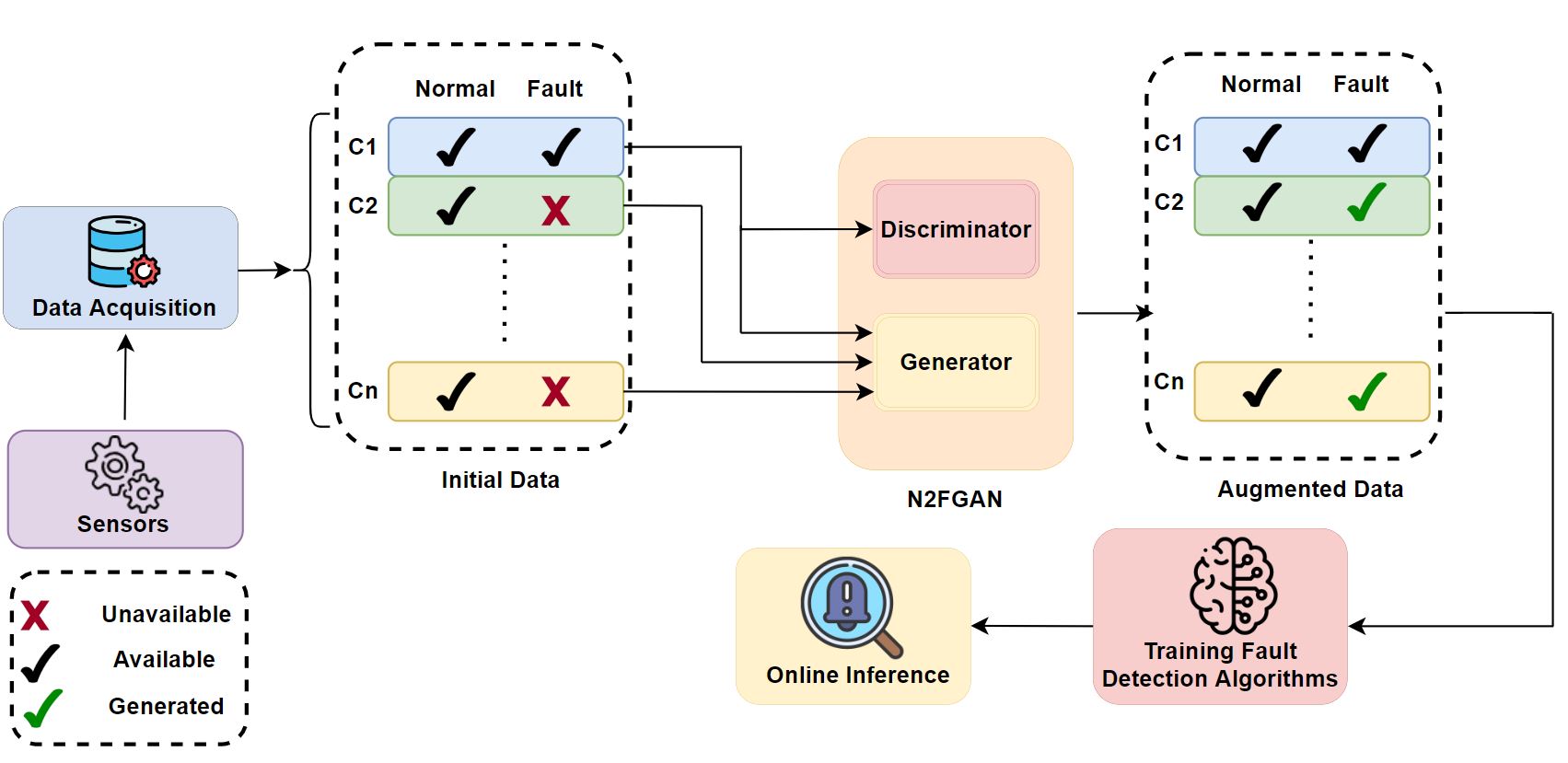}
\caption{A flow diagram of using N2FGAN for fault data generation.\label{fig1}}
\end{figure}

 In this paper, a novel method inspired by image to image translation \cite{s30} is introduced and tested on vibration signals to generate fault data from normal data. Pairs of normal and fault data are fed as inputs to the network at a given condition. After the training phase, the network can generate new fault data under different conditions. It has been assumed no fault sample is available in other conditions, and this data generation is only done using normal data. The efficiency of the proposed method and the quality of generated data are evaluated using different classifiers and visualization methods. The paper's organization is as follows: A literature review is done in Section 2. Next, Section 3 introduces some background theories and Section 4 elaborates on the proposed method, normal to fault GAN (N2FGAN). The N2FGAN is tested on the CWRU (Case Western Reserve University) dataset and verified in Section 5, and finally, conclusions are available in Section 6. In figure~\ref{fig1} a flow diagram of using N2FGAN to generate synthetic fault data for new conditions in different systems is shown. $c_1$,$c_2$,..., $c_n$ refers to different working conditions.

\section{Literature Review}
Numerous AI techniques, such as traditional machine learning methods and deep learning approaches, have been used for fault recognition and diagnosis in roller ball bearings or rotating parts of machinery. Lei et al. \cite{s1} reviewed systematically the development of Intelligent Fault Diagnosis (IFD) since the adoption of machine learning approaches, presenting the past, present, and future Artificial Intelligent approaches. Schwendemann et al. \cite{m20} surveyed machine learning in predictive maintenance and condition monitoring of bearings and studied different approaches to classify bearing faults and the severity detection.
\par Liu et al. \cite{s2} also presented a review of literature on the applications of Artificial Intelligence algorithms for the fault detection of rotating machinery with a focus on traditional machine learning methods such as Naive Bayes classifier, K-Nearest Neighbours (k-NN), Support Vector Machines (SVM), and Artificial Neural Network algorithms. In earlier years of intelligent fault diagnosis, traditional machine learning approaches involved collecting raw sensor data of various fault types, extracting features from the collected data, and developing diagnosis models from the features to automatically recognize machines' health status. Although traditional machine learning methods can automate the fault detection process, these approaches cannot handle increasingly large data due to their low generalization performance, thereby reducing their accuracy in fault diagnosis. For instance, support vector machine classifiers can be applied to classification and regression problems. However, they do not perform well when applied to multi-class classifications or pair-wise classification problems. Some approaches like SVMs are computationally expensive  and can not deal massive industrial data efficiently \cite{s3}.
 \par In recent times, deep learning paradigms for intelligent fault diagnosis have become prominent because they can automatically learn fault characteristics from the data without direct feature extraction. Also, they can handle large amount of industrial data which is one of the drawbacks of traditional machine learning methods and this has helped to reform intelligent fault diagnosis since the 2010s. Li et al. \cite{s4} reviewed the literature on the applications of deep learning methods for fault diagnoses, analyzing the deep learning approaches in relevant publications to point out the advantages, disadvantages, areas of imperfections, and directions for future research.
Although the adoption of deep learning methods has led to many successes, these approaches assume that labeled data is sufficient for training diagnosis models \cite{s1}. However, this assumption is impractical given the working conditions in most industries. The collected data is inadequate as machines seldom develop faults, and more healthy condition data are collected than faults. So, even with deep learning approaches, the collected data are unbalanced and insufficient to train reliable fault diagnosis models. This poses some limitations of using intelligent fault diagnosis in industries.
As mentioned earlier, the lack of fault data during the network training process is generally termed as a small sample problem \cite{s5}. Researchers have come up with three significant ways of solving the small sample problem, which are data augmentation-based, transfer learning/domain adaptation-based, and model-based strategies. The data augmentation and transfer learning-based methods attempt to increase the amount of data by generating similar data from the existing fault data; slightly modified copies of existing fault data are used to create synthetic data for training the neural network, transfer learning-based approaches use pre-trained networks from similar domains to train the new models in a bid to minimize the amount of data required for training.
\par GANs have unveiled promising capabilities in intelligent fault diagnosis for data argumentation and adversarial training purposes. They can be considered as a potential solution to the small sample problem because GANs can be used to generate additional data with the same distribution as the original data. The Generative Adversarial Network was first introduced by Goodfellow in 2014 \cite{s7}. Generally, a standard GAN comprises two modules, the generator and the discriminator. The generator learns the distribution of training data, and a discriminator's goal is to distinguish the samples of the original training set from the generated ones.
This capability exhibited by the Generative Adversarial Network has made its application in intelligent fault diagnosis vast. Pan et al. \cite{s6} reviewed the related literature on small-sample-focused fault diagnosis methods using GANs. Their paper describes the GAN approaches and reviews GAN-based Intelligent Fault Diagnosis applications in literature while discussing the limitations and future road maps of GAN-based fault diagnosis applications. Li et al. \cite{s8} also presented research on GANs with a focus on the theoretical development and achievements of GANs while introducing and discussing the improved GAN methods and their variants.
\par 
% The  are different types of GAN-based data augmentation methods based on the dimensions of the original signal: one-dimensional time-domain signals, one-dimensional frequency domain signals, two-dimensional image signals, and one-dimensional feature sets \cite{s6}.
Liu et al. \cite{s8} presented a rotating machinery fault diagnostics framework that is based on GANs and multi-sensor data fusion to generate synthetic data from the original data. Zhang et al. \cite{s9}, Wang et al. \cite{s10}, and Lv et al. \cite{s11} all made use of one-dimensional time-domain signals to generate synthetic data using GANs for classification and diagnosis of rotating machinery. Similarly, Li et al. \cite{s12}, Wang et al. \cite{s13}, Zheng et al. \cite{s14}, and Wang et al. \cite{s15} used one-dimensional frequency domain signals, and Huang et al. \cite{s16} and Shi et al. \cite{s17} used two-dimensional images while Pan et al. \cite{s18}, and Zhou et al. \cite{s19} used one-dimensional feature sets to generate synthetic data.
\par The original GAN has been extended into various forms such as the Wasserstein GAN (WGAN),  Convolutional-based GANs, Semi-supervised GANs, and Condition-based GANs  to enhance the quality of data synthesis and improve the training process. For instance, the complexities of controlling the adversarial process between generator and discriminator cause a mode collapse/gradient disappearance phenomenon leading to unsatisfactory data generation performance of the GAN models. To overcome this challenge, Arjovsky et al. \cite{s20} introduced the Wasserstein GAN to deal with mode collapse phenomena. It provided a solution to the instability problem of GAN but had the challenge of weight clipping which was addressed by Gao et al. \cite{s21} through the combination of WGAN with a gradient penalty. Zhang et al. \cite{s22} also tried solving the small sample problem, focusing on intelligent fault diagnosis via multi-module gradient penalized GAN. The proposed method comprises three network modules: generator, discriminator, and classifier. The mechanical signals are generated by adversarial training and are then used as training data. \cite{s23}, and \cite{s24} also used GANs for the fault diagnosis problem of rotating machinery.
\par These improved variants of GANs have been extensively applied to roller-bearing fault diagnosis. There have also been many combinations of GANs with other generative models for fault diagnosis, namely encoder, auto-encoder (AE), and variational auto-encoder. Wang et al. \cite{s13} combined GAN and conditional variational auto-encoder to enhance the quality of generated samples for fault pattern recognition in planetary gearboxes. \cite{s25} proposed an improved fault diagnosis approach to learn the deep features of the data by combining an encoder with GANs and integrating the discriminator with the deep regret analysis method to avoid mode collapse by imposing the gradient penalty on it.
\cite{s25} also proposed a novel method called upgraded GAN, which is a combination of Energy-based GANs, Auxiliary-classifier, and conditional variational autoencoders. Some other applications of GANs for data argumentation in literature for fault diagnosis are demonstrated by Liu et al. \cite{s27}, who proposed a data synthesis approach using deep feature enhanced GANs for roller bearing fault diagnosis, and \cite{s28} which used wavelength transform to extract image features from time-domain signals with GANs for generating more training samples and CNN for fault detection.
Generative Algorithms have proven to be beneficial for solving the small sample problem encountered when using data-driven approaches for Intelligent fault diagnosis. This method is widely accepted, and more improvements and modifications to the standard GAN have been embraced in literature to develop highly effective models capable of detecting and classifying industrial fault data and other applications in intelligent fault diagnosis; it is also adopted in this research work to develop new fault samples.

\section{Background}
\par  This section provides a brief introduction to the networks and algorithms that are used in the paper. Long Short-Term Memory (LSTM) network is an important concept and used for classification, and the CNN network is used in both the architecture of the data generation algorithm and the classifier. Conditional GAN is the base of N2FGAN. Understanding its architecture is essential for comprehending the image-to-image translation and N2FGAN. 
\subsection{LSTM}
\par LSTM is a type of Recurrent Neural Network (RNN) and is one of the most potent classifiers in machine learning. The network's efficiency and impressive ability stem from the formulation of the network and its learning algorithms.
%; LSTMs are used to model time prediction tasks, so we refer to them as dynamic classifiers. RNNs are used mostly with sequential data and time-series data and are commonly used for problems such as Language Translations, Natural language processing, and prediction problems where
The output of a network is influenced by the information from the previous and current inputs. RNNs are an extension of the feed-forward neural networks but are distinguished by their memory.
\par RNNs are dynamic systems \cite{t1} with an internal state at each time step of the classification resulting from the connections between higher layer neurons and the neurons in the lower layers as well as optional self-feedback connection(s).
% These feedback connections enable RNNs to propagate data from earlier events to current processing steps, thus, allowing RNNs to build a memory of time-series events. 
Initially developed RN networks such as Elman and Jordan networks \cite{t2} had a limitation of looking back in time for more extended time steps due to the issue of vanishing or exploding gradients. Long Short-Term Memory Recurrent Neural Networks were developed to address this issue.
% LSTM networks help to bridge minimal time lags for more than 1,000 time steps, depending on the complexity of the built network \cite{t3} with constant error carousels (CECs) \cite{t2}. Constant error carousels enforce a constant error flow within special cells, enabling the RNNs to remember previous inputs for a long time. Multiple gates handle access to these cells, which learn when to grant access to constant error flow through the CESs. 
LSTMs usually have three gates, the input, forget, and output gates that learn overtime what information(s) are essential; the input gates determine whether a piece of information is important and usually use simple sigmoid function units with activation range between 0 and 1 to control the signal into the gate. The forget gate helps to decide if a piece of information should be deleted or kept, while the output gate learns how to control access to cell contents and helps to decide which information is worthy of impacting the output of the current time-step.

\begin{figure}
\includegraphics[width=12.5cm]{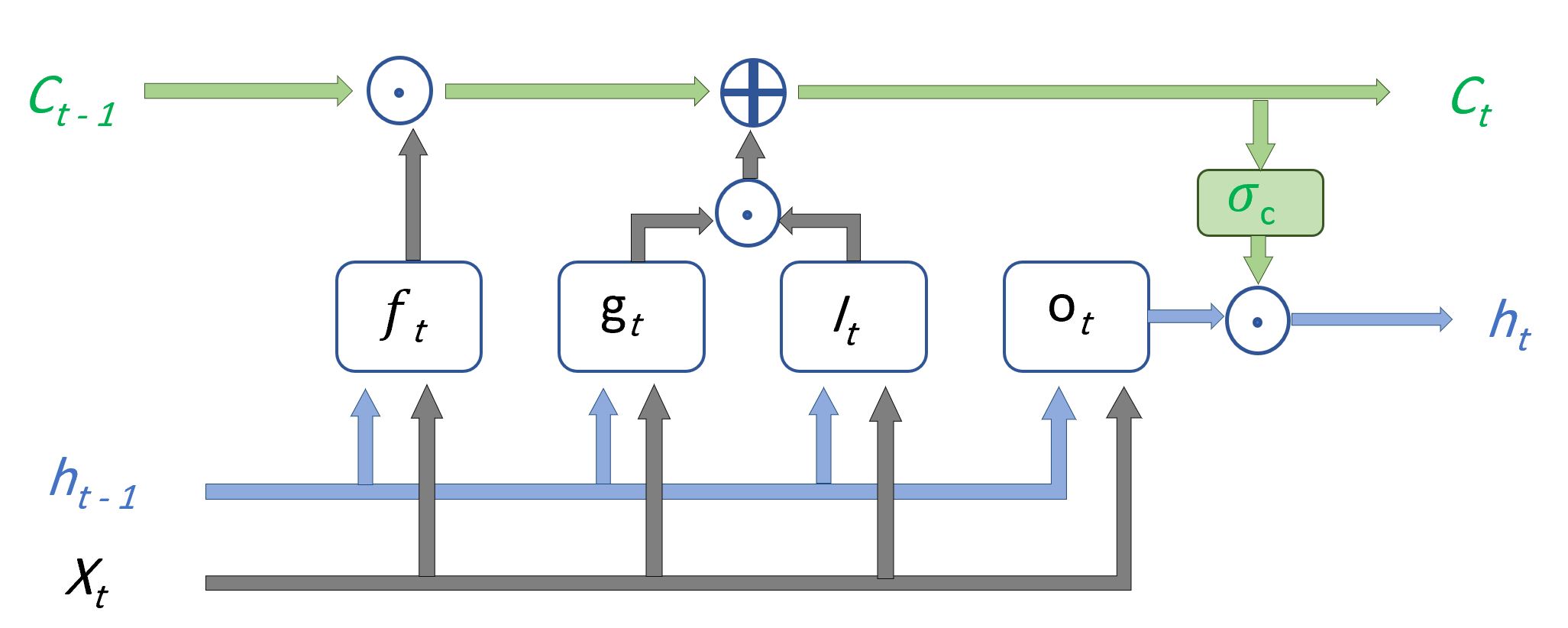}
\centering
\caption{A basic structure of LSTM Network.\label{fig2}}
\end{figure}

Figure~\ref{fig2} is the basic structure of LSTM-RNN; where:
$f_t$ is forget gate, $g_t$ is cell candidate, $i_t$ is input gate, $o_t$ is output gate, $C_t$ is Cell state, and $h_t$ is hidden state.

\subsection{CNN}
\par Convolutional Neural Networks (CNN) are artificial neural networks that are primarily used to solve image-driven pattern recognition tasks because of the design and structure of their architecture. They have been widely used in fault detection \cite{shao2022unsupervised, tang2021novel,kahr2022condition} because of its feature extraction capability. The idea behind convolutions is to use kernels to extract particular features from input data.
% Traditionally, experts design kernels for specific feature extraction tasks such as edge detection and image sharpening but in convolutional neural networks, the kernels are learned in the network.
\par
\begin{figure}
\centering
\includegraphics[width=10cm]{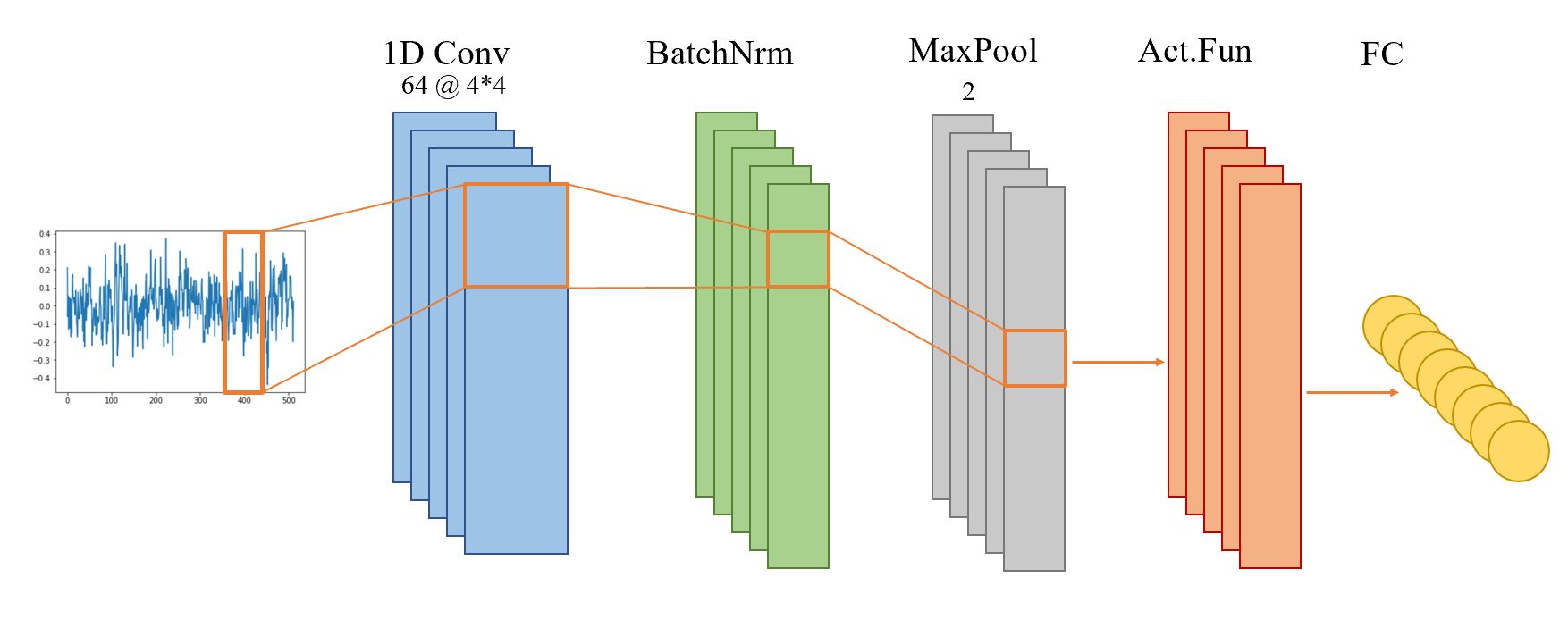}
\caption{A schematic representation of the major CNN layers.\label{fig3}}
\end{figure}

 CNNs are composed of 3 main layers: the convolutional layers, pooling layers, and fully connected layers. The convolutional layer parameters use learnable kernels \cite{s29}. These kernels, usually 2-dimensional for image recognition tasks or 1-dimensional for time series data, glide over the entire depth of the input while calculating the scalar product for each kernel.
% The network can learn kernels that detect specific features at a given spatial position of the input by combining the pixels from a small area of the input data, that is, combining pixels within the same locality.  This is particularly useful because pixels usually appear in a consistent order, and a pixel influences the surrounding pixels.If there are deviations, this anomaly could be converted into a feature, and all this can be detected by comparing a pixel with other pixels in its locality. 
Then an activation function is used to enhance the nonlinear expression of the convoluted features. The process is shown in Equation~\ref{eq:1}., where $x$ is the signal, $f_k$ is kernel filter, $b_k$ is a bias, and $\sigma$ is activation function. The pooling layers reduce the representation's dimensionality, reducing the model's computational complexity and allowing for better generalization. The most common pooling method in CNN is max-pooling $max$, which calculates the maximum value in a range $w$, as shown in Equation~\ref{eq:2}. The fully connected layer consists of neurons directly connected to the neurons in the two adjacent layers, just like traditional ANN. Figure \ref{fig3} is a schematic representation of CNN with convolution, batch normalization, pooling, activation function, and fully-connected layers \cite{m18}.

\begin{equation}\label{eq:1}
h_k = \sigma (x*f_k+b_k)
\end{equation}

\begin{equation}\label{eq:2}
h_pk = max (h_k,w)
\end{equation}

\subsection{Conditional GAN}
\par Generative Adversarial Networks consist of two networks trained simultaneously: (1) a generative model, $G$, capturing the data distribution to produce synthetic samples $\tilde x=G(z)$ while the input of the network is a random noise vector $z$, with the distribution of $\mathscr{P}_z$; (2) a discriminator model, $D$, which discovers if the sample is generated by $G$ or is a real sample from the training data. $G$ is trained in a way to deceive $D$ by making convincing, realistic samples. On the other hand, $D$ estimates the corresponding probability of each sample to find out the source. The value function of GAN is defined in Equation~\ref{eq:3}.
\begin{equation}\label{eq:3}
\begin{split}
\min_{G}\max_{D}V(D,G)= \mathbb{E}_{x\sim \mathscr{P}_{r}}[\log(D(x))] + \mathbb{E}_{\tilde x \sim \mathscr{P}_{f}}[\log(1-D(\tilde x))],
\end{split}
\end{equation}

where $\mathscr{P}_{r}$ and $\mathscr{P}_{f}$ denote the distribution of the raw data and of the synthetic samples,    respectively \cite{s7}.

% \begin{figure}
%     \centering
%     % \includegraphics[width=8.2cm]{figs/p2_GAN.PNG}
%     \caption{The schematic process in GANs}
%     \label{fig:gan}
% \end{figure}
	
% \subsection{Conditional GAN}
\par Conditional Generative Adversarial Network is a variation of GAN. It places a condition on generator and discriminator by feeding some extra information, $y$. This information could be data from other modalities. $y$ is fed into both generator and discriminator\cite{m12}. The objective of the CGAN is expressed in Equation~\ref{eq:4}. 
\begin{equation}\label{eq:4}
\begin{split}
\min_{G}\max_{D}V(D,G)= \mathbb{E}_{x\sim \mathscr{P}_{r}}[\log(D(x|y))] + \mathbb{E}_{\tilde x \sim \mathscr{P}_{z}}[\log(1-D(G(z|y))],
\end{split}
\end{equation}

\subsection{Image to Image Translation}
\par The proposed method used in this paper was inspired by the Pix2Pix, an image-to-image translation algorithm introduced by Phillip Isola et al. \cite{s30}. Image to image translation is transforming an image from one domain to another.  It is mapping an input image and an output image, mapping a day image to a night as an instance. Pix2Pix is the pseudonym for implementing a generic image-to-image translation solution that involves mapping pixels to pixels using CGANs. As mentioned earlier, GANs map a random noise vector to an output, while CGANs learn a mapping from an observation and random noise vector to the output. The framework used in \cite{s30} differs from other CGANs frameworks because it was designed not to be application-specific like image-to-image translation methods. %It is generic in the sense that a Pix2Pix network trained on pairs of related images, say "A" and "B", learns to convert an image of type "A" into an image of type "B", or vice versa. So, two different networks need not be trained to convert from Image "A" to "B" and from "B" to "A".
The main characteristics of Pix2Pix compared to CGAN is that its generator's input does not include random noise. This would make the output of the generator deterministic. To address this, the noise is added in the form of dropout layers to the generator's architecture.
Moreover, Pix2Pix chooses different architectures for its generator and discriminator, where U-Net, a convolutional network using for image segmentation relies on data augmentation to use available samples more effectively\cite{m21}. and PatchGAN \cite{m22} are used, respectively. Both use modules of the form convolution-BatchNorm-ReLu.
\par In our proposed work, a modification of Pix2Pix  was adopted on vibration signals because of its generic nature and ability to work well on problems framed as an image-to-image translation.

\section{Proposed Model (N2FGAN)}
\par
\begin{figure}
\centering
\includegraphics[width=15cm]{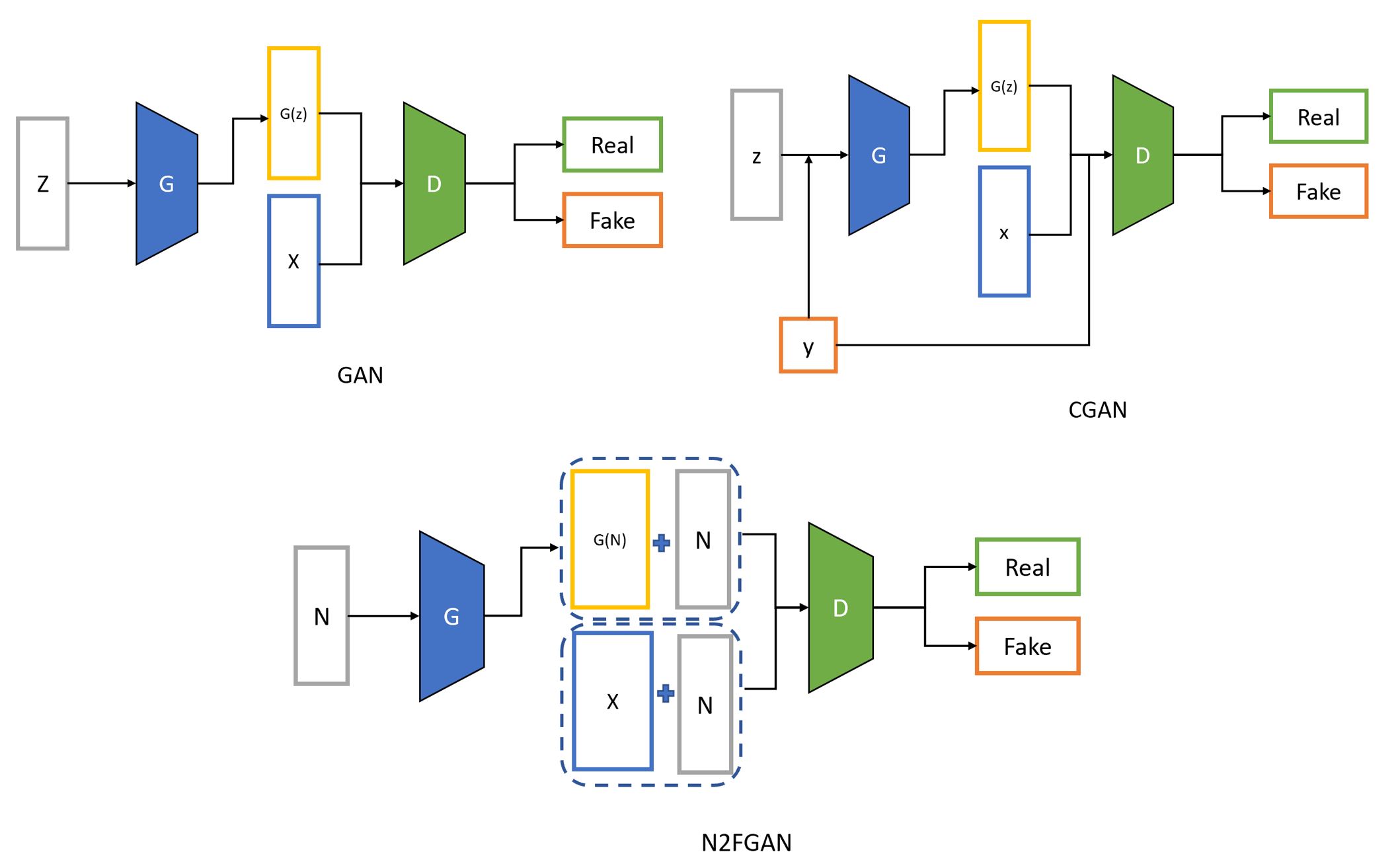}
\caption{A comparison between GAN, CGAN, and our proposed methods(N2FGAN).\label{fig4}}
\end{figure}

 As mentioned before, fault data is scarce in industrial plants; however, normal data is ample. This section will introduce a novel method for generating fault data from normal data. This method is inspired by image-to-image translation, which is a variation of CGAN for the conditional data generation task. A conventional GAN learns a mapping from random noise vector $z$ to the output synthetic sample $\tilde{x}$ = $G(z)$. However, CGANs learn to map from observed information $y$ and random noise vector $z$ to $\tilde{x}$.The CGAN model has been improved, making it suitable for signal-to-signal translation. In our proposed method (N2FGAN), similar to Pix2Pix framework, normal data is used as the input of the generator without any random noise. In Figure \ref{fig4}, a comparison between GAN, CGAN and N2FGAN is shown.

\subsection{Network Architectures}
\subsubsection{Generator}
\par An encoder decoder structure is used for the generator. The input is passed through some down sampling layers known as encoder, and then the process is reversed in the decoder. In this process, hidden information of the data is extracted. The generator consists of an encoder and a decoder. Each block in the encoder consists of a 1D convolutional layer, a Batch Normalization layer, and a Leaky ReLU layer. Each block in the decoder consists of Transposed convolution, Batch Normalization, dropout, and Relu layers. There are skip connections between the encoder and decoder. The generator consists of four blocks in encoder and decoder, where the input data is a vector of length 512, and the dimension of the latent space is 64.
\subsubsection{Discriminator}
\par The discriminator is a convolutional classifier. It includes three convolutional blocks, each consisting of a 1D convolutional layer, a Batch Normalization layer, and a Leaky ReLU Layer. It receives two pairs of concatenated inputs, the first input is the normal data and the actual fault data, from the dataset that should be classified as real data, and the second input is the concatenated normal data and generated fault data from the generator's output, which is synthetic.
\subsection{Objective}
\par The total loss consists of the generator and the discriminator losses. The generator loss is a sigmoid cross-entropy loss of the generated data and an array of ones. To make generated data structurally similar to the target data, the $L_1$ distance is used. L1 loss is defined as mean absolute error between the generated data and the target data.
\begin{equation}\label{eq:5}
\begin{split}
\mathcal{L}_{L1}(G)= \mathbb{E}_{x,\tilde{x},z}[||\tilde{x}-G(x,z)||_{1}],
\end{split}
\end{equation}
The discriminator loss consists of the sum of sigmoid cross-entropy loss of the actual data and array of ones and sigmoid cross-entropy loss of the generated data and array of zeros, respectively. The final objective of the proposed method is defined as follows, where $\lambda$ is a constant and considered 100.
\begin{equation}\label{eq:6}
\begin{split}
{G^*} = \min_{G}\max_{D}V(D,G) + \lambda*\mathcal{L}_{L1}(G),
\end{split}
\end{equation}
\section{Experiments and Discussion}

\begin{figure}
\centering
\includegraphics[scale=.75]{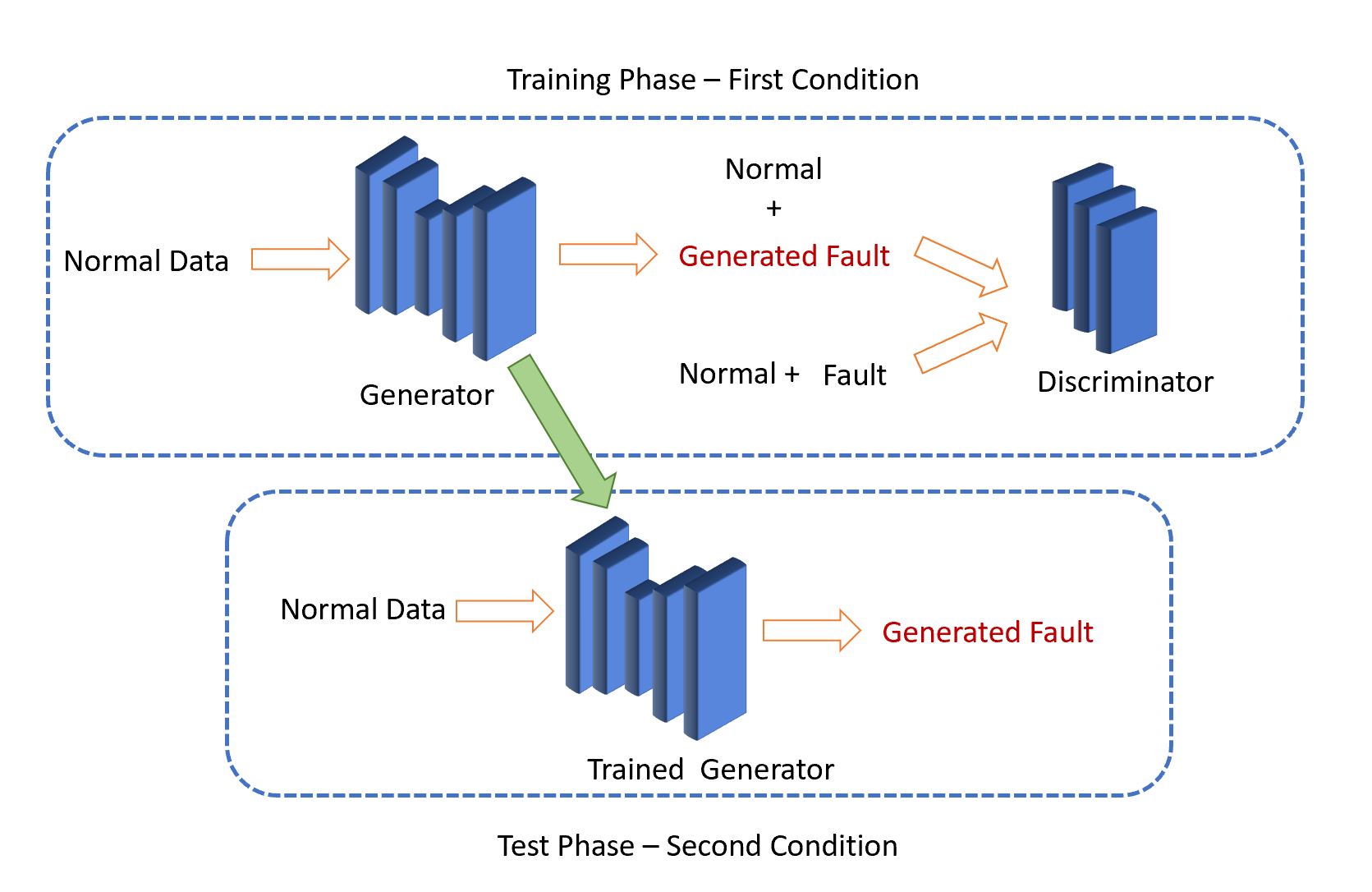}
\caption{The propose data generation framework (N2FGAN).\label{fig5}}
\end{figure}

In the following, N2FGAN is implemented for generating new fault data. The network is trained in the first condition, where both normal and fault data are available. In the next step, the trained generator, makes fault data from normal in a new condition where there is no fault data available. The model is tested on a real-world dataset in various conditions. Finally, three classifiers are used for the evaluation of the generated data. In this case, the machinery motor loads and motor speeds are considered as different conditions. Some statistical features of the generated data are compared with the actual data, and some visualizations are also created to show the quality of the generated data. In this paper, actual data is referred to as the samples from the dataset, and generated data is the output of the generator network. An overview and steps of the proposed method are shown in Figure~\ref{fig5}.  All experiments were performed using Python 3.7 on a computer with a GPU of NVIDIA Tesla P100 and 16 GB of memory.

% All experiments were done using Python 3.7 on a computer with a GPU of NVIDIA Geforce GTX 950 and 16 GB of memory.
\subsection{Dataset Description}
\par
\begin{figure}
\includegraphics[width=0.6\textwidth]{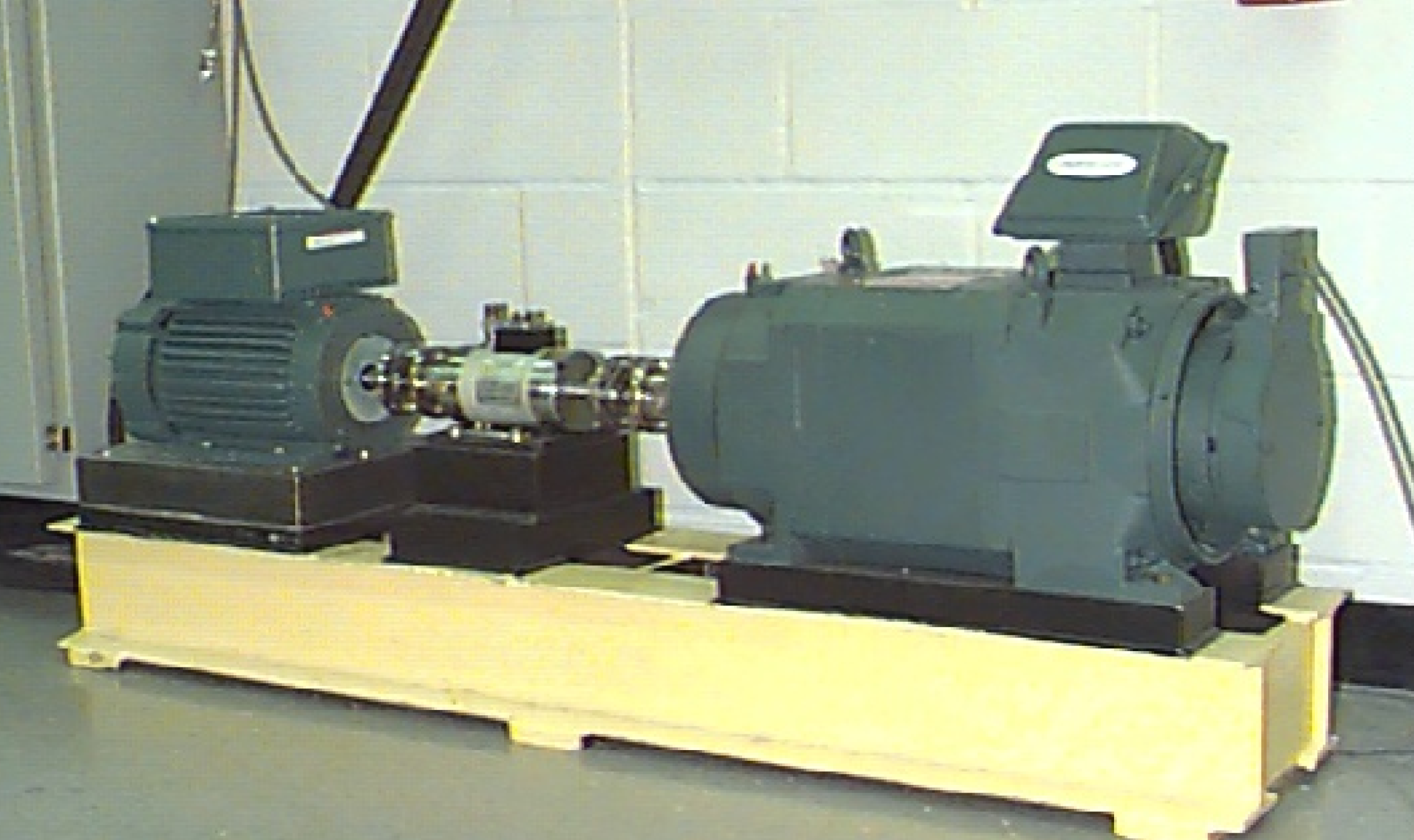}
\centering
\caption{CWRU bearing data collection test bed.\label{fig6}}
\end{figure}

  In this paper, CWRU bearing dataset (\url{https://engineering.case.edu/bearingdatacenter} (accessed on 20 June 2022)). is used for data generation. The testbed is shown in Figure~\ref{fig6}.  The data was collected using an experimental setup consisting of a 2 hp Reliance Electric motor, a torque transducer/encoder, a dynamometer, and control electronics. Acceleration data was collected at 12,000 samples/second from fan-end and drive-end of the machine. Faults are made artificially using electro-discharge machining (EDM), and the diameter of faults range from 0.007 inches to 0.040 inches. There are three fault categories, inner raceway, rolling element, and outer raceway. Outer race fault is collected in three different orientations: directly in the load zone, orthogonal to the load zone and opposite to the load zone, so there are five different type of faults in total. Vibration data was recorded for motor loads between 0 and 3 horsepower with speeds of 1730 to 1797 RPM. In the experiment, drive-end data is used. To generate different datasets signal burst of length 200 is used, and 100 SNR noise is added to make data generation more complex.

\subsection{Training phase of the data generation algorithm}
\par
\begin{figure}
	\centering
		\includegraphics[width=\textwidth]{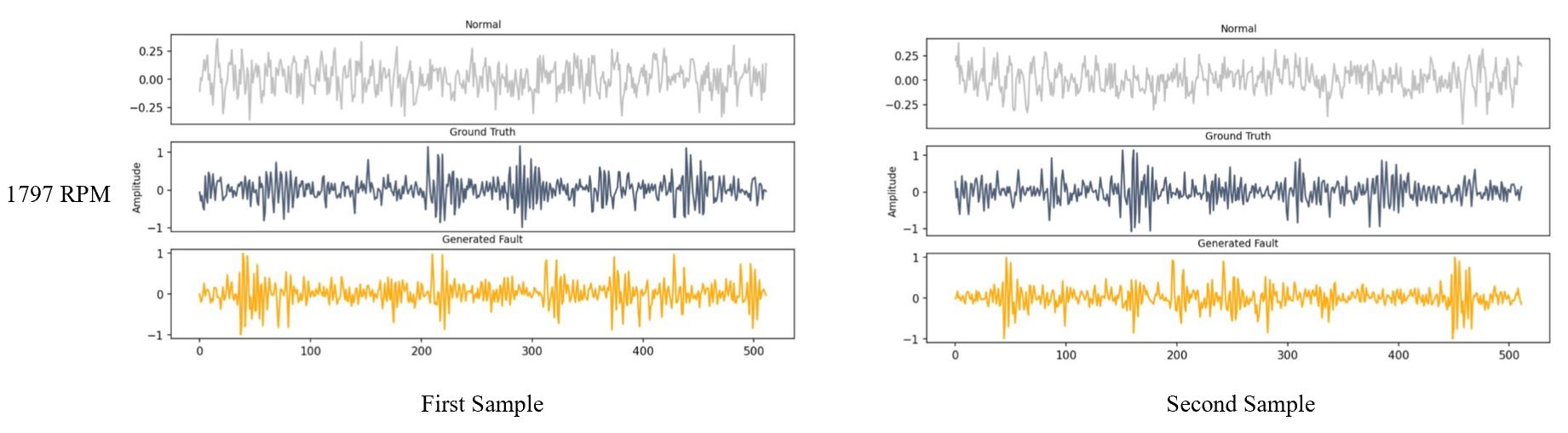}
	\caption{Some samples of generated fault data in 1797 RPM.}
	\label{fig7}
\end{figure}

 The defined architecture is used to generate fault data in the first condition (in this case first RPM), where both normal and fault data are available. The process is done by applying Adam optimizer with a learning rate of 0.0002 over 4000 steps. The experiment is done on actual normal and inner race faults with 0.007 inches diameter in the 1797 RPM. In this case, although the fault data is available, more fault data is generated. Some samples of the normal data, actual fault data, also known as ground truth, and generated fault data in the training phase are shown in Figure~\ref{fig7}.

\subsection{Data generation in new condition}
\par
\begin{figure}
	\centering
		\includegraphics[width=\textwidth]{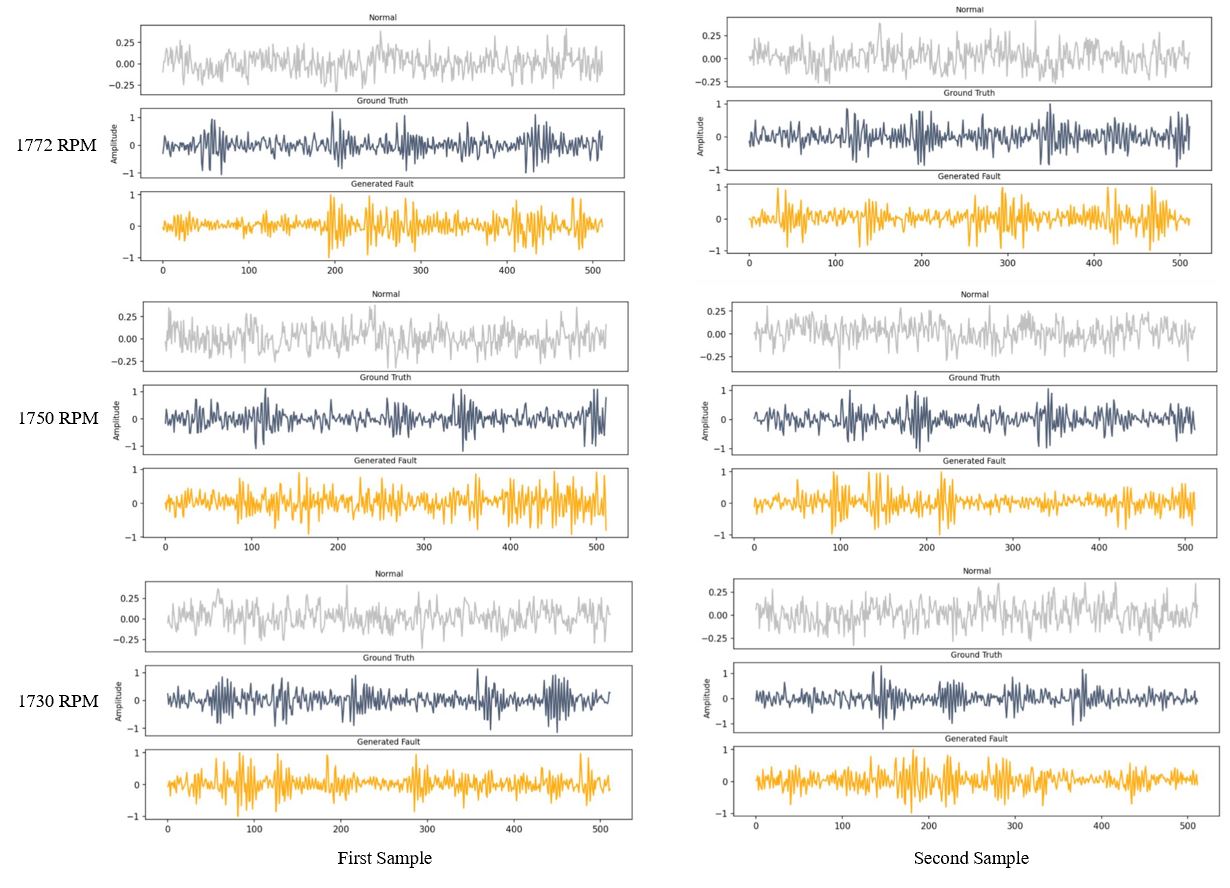}
	\caption{Some samples of generated fault data for different RPMs.}
	\label{fig8}
\end{figure}

 The trained generator is used for generating fault data in new conditions; these conditions are defined as different working speeds at 1772, 1750, and 1730 RPM, where it has been assumed there is no sample of fault data; however, the generated fault data is compared to the actual data for evaluation. The generator's input is normal data in new conditions, and the fault data is not used in the data generation process. Some samples of the generated data in the different conditions using the trained network are shown in Figure ~\ref{fig8}. The generator is trained in 1797 RPM, and the same network is used for data generation in all conditions.

\subsection{Evaluation}
\par
\begin{figure}
\centering
\includegraphics[width= 17cm]{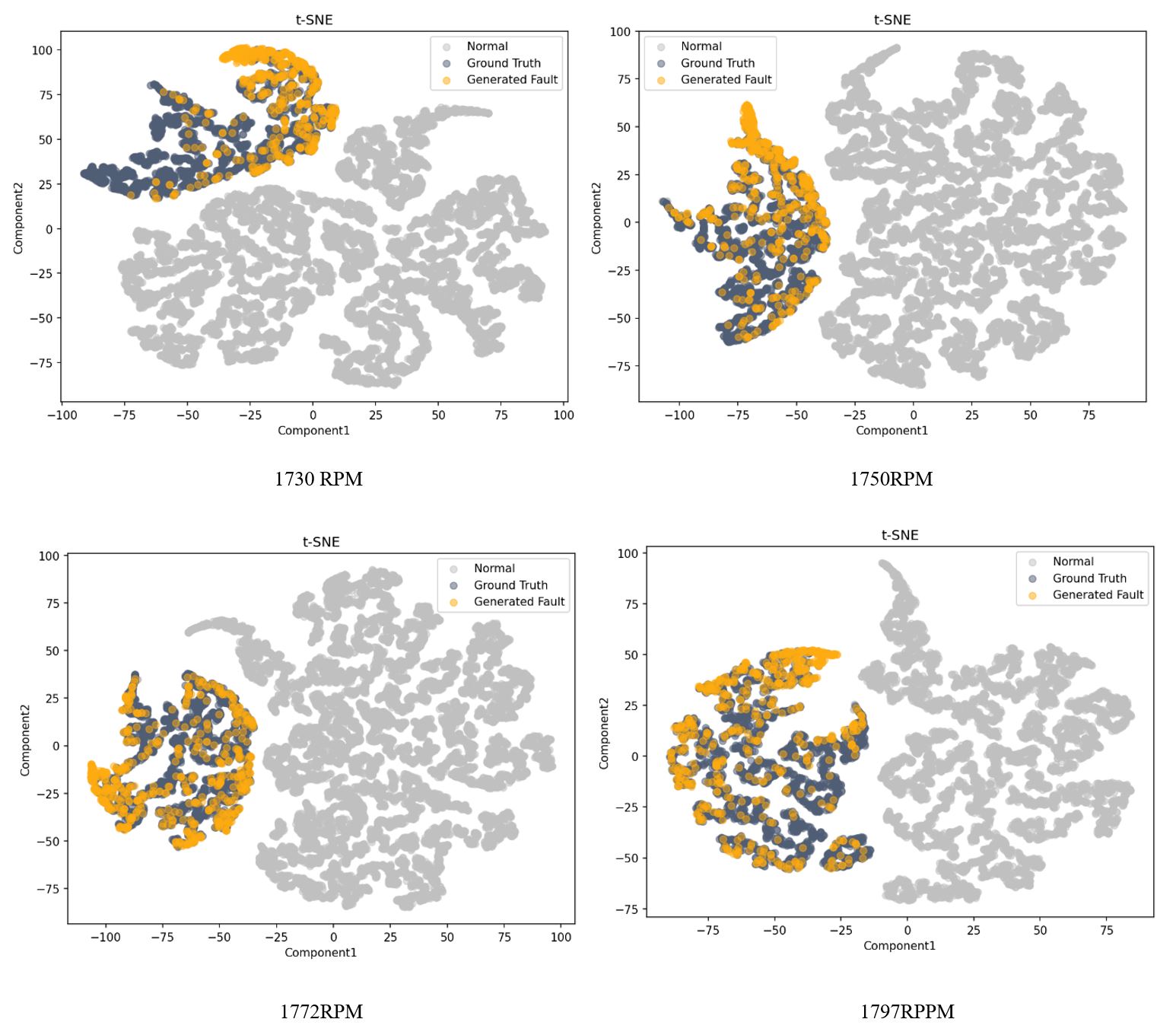}
\caption{t-SNE visualization of generated data\label{fig9}}
\end{figure}

 Evaluating the quality of the generated data is a difficult task. The generated samples should be similar to the actual data in any conditions. Neural network-based classifiers are used to validate the generated data alongside a statistical comparison with t-SNE (t-distributed stochastic neighbor embedding) to represent the statistical distributions. Both time domain and frequency domain features are extracted based on work in \cite{yu2015novel}, and used for the t-SNE visualization. The selected features are listed in Table \ref{tab1} and the 2-component t-SNE plots for different conditions are shown in Figure \ref{fig9}. The generated fault data is easily separable from normal condition data, and it is very similar to the ground truth which shows that the distributions of the generated signal features closely match the distributions of actual fault features.

\begin{table} 
\caption{Selected features for analysis of generated fault data.\label{tab1}}
\begin{adjustbox}{width=\columnwidth,center}
\begin{tabular}{ccccc}
\toprule
\textbf{Time Domain Feature}	& \textbf{Formula} & \textbf{Frequency Domain Feature}  & \textbf{Formula} \\
\midrule
Mean               & $\bar{x} = \frac{1}{N}\sum_{i=1}^{N}x(i)$                                 & Mean               & $\bar{f} = \frac{1}{N}\sum_{i=1}^{N}f(i)$                                      \\
Standard Deviation & $\sigma = \sqrt{\frac{1}{N-1}\sum_{i=1}^{N}(x(i)-\bar{x})}$               & Standard Deviation & $\sigma_f = \sqrt{\frac{1}{N-1}\sum_{i=1}^{N}(f(i)-\bar{f})}$                  \\
Skewness           & $\tilde{\mu}_3 = \frac{\sum_{i=1}^{N}{(x(i)-\bar{x})^3}}{(N-1)*\sigma^3}$ & Skewness           & $\tilde{\mu}_{3f} = \frac{\sum_{i=1}^{N}{(x(i)-\bar{f})^3}}{(N-1)*\sigma_f^3}$ \\
Crest Factor       & $CF = \frac{max|x(i)|}{\sqrt{\frac{1}{N}\sum_{i=1}^{N}{x(i)^2}}}$         & Crest Factor       & $CF_f = \frac{max|f(i)|}{\sqrt{\frac{1}{N}\sum_{i=1}^{N}{f(i)^2}}}$            \\
Kurtosis           & $\kappa = \frac{1}{N}\sum_{i=1}^{N}{x(i)-max(x(i))^4}$                    & Shannon Entropy              & $-\sum_{i=1}^{N}{f(i)log(f(i))}$                                               \\ 
\bottomrule

\end{tabular}
\end{adjustbox}
\end{table}

\par In the first step, a Binary LSTM classifier with a Softmax layer is used to determine if the generated data is faulty or normal\cite{m16}. 
% In this approach, a concern is that the input data may not belong to any of both categories. To address this problem, we consider a limit for the softmax output. If the probability of belonging to the faulty class was above 0.65, we could classify the data as fault data.
The classifier was trained on the actual dataset. After the training phase, the actual data of the target class was replaced with the generated data samples and fed into the classifier as test data. The test data consists of normal, inner race, ball, the outer race centered, outer race orthogonal, and outer race opposite fault data with 0 to 5 labels, respectively, and there are 480 samples in each class. The experiment was done several times on binary and multiclass classifiers. The binary classifiers' accuracy for generated data in the same condition and the new condition is 100\%, which shows that the generated fault data is not similar to normal. Three multiclass classifiers,  convolutional LSTM (ConvLSTM), CNN and convolutional auto-encoder (ConvAE), were used in this experiment to evaluate the effectiveness of adding synthetic samples. Their details are shown in the table \ref{tab2}.
% As mentioned before, the dataset has five different faults and one normal class. 
The multi-class classifiers also illustrate a high performance, having more than 97\% of accuracy. Table~\ref{tab3} shows the performance of the classifiers applied to the generated data.

\begin{table} 
\caption{Classifiers description.\label{tab2}}
\newcolumntype{C}{>{\centering\arraybackslash}X}
\begin{tabularx}{\textwidth}{p{2cm}p{11cm}}
\toprule
\textbf{Framework}	& \textbf{Description}\\
\midrule
 ConvLSTM & The architecture consists of two CNN blocks (containing
1D-Convolutional layers, Batch Normalization, ReLU, and Max Pooling,), an LSTM block, a Dense layer with Sigmoid activation function, a Dropout, and a SoftMax layer.\\ 
\midrule
 CNN & It consists of four CNN blocks (containing one 1D-Convolutional layer, Batch Normalization, ReLU, and Max Pooling layer), A flatten layer, a fully connected layer, and a SoftMax classification layer. \\
 \midrule
 ConvAE & It is a multi-layer network consisting of an encoder and a decoder. Each includes three CNN blocks (containing 1D-Convolutional layers, ReLU and Max Pooling or upsampling), A flatten, a fully connected layer, and a SoftMax classification layer. \\
\bottomrule
\end{tabularx}
\end{table}
\unskip

% \begin{table} 
% \caption{Classifiers accuracy for different condition while the training condition is 1797 RPM.\label{tab3}}
% \newcolumntype{C}{>{\centering\arraybackslash}X}
% \begin{tabularx}{\textwidth}{CCCC}
% \toprule
% \textbf{Condition}	& \textbf{ConvLSTM classifier}	& \textbf{CNN classifier}& \textbf{ConvAE classifier}\\
% \midrule
%  Test 1797 & 99.67\% & 98.34 \%  & 99.33 \%\\ 
%  Test 1772 & 98.83\% & 98.01  \% & 96.19 \% \\
%  Test 1750 & 99.17\% & 97.68\% & 98.67 \%  \\
%  Test 1730 & 98.34 \% & 99.01\% & 97.01 \%\\
% \bottomrule
% \end{tabularx}
% \end{table}
% \unskip

\begin{table} 
\caption{Classifiers accuracy, $F_1$ score, precision and recall for test data in different conditions while the training condition is 1797 RPM.\label{tab3}}
\begin{adjustbox}{width=\columnwidth,center}
\newcolumntype{C}{>{\centering\arraybackslash}X}
\resizebox{0.9\paperwidth}{!}{\begin{tabular}{ccccccccccccc}
\hline
Condition & \multicolumn{4}{c}{ConvLSTM Classifer}   & \multicolumn{4}{c}{CNN Classifer}        & \multicolumn{4}{c}{ConvAE Classifer}     \\ \cline{2-13} 
          & Accuracy & $F_1$ score & precision & Recall & Accuracy & $F_1$ score & Precision & Recall & Accuracy & $F_1$ score & Precision & Recall \\ \hline
1797      & 98.89\%    & 98.89\%    & 98.9\%      & 98.89\%  & 99.34\%    & 99.34\%    & 99.35\%     & 99.34\%  & 99.38\%    & 99.37\%    & 99.38\%     & 99.37\%  \\
1772      & 98.78\%    & 98.78\%    & 98.81\%     & 98.78\%  & 98.85\%    & 98.85\%    & 98.89\%     & 98.85\%  & 99.27\%    & 99.70\%     & 99.28\%     & 99.27\%  \\
1750      & 99.24\%    & 99.24\%    & 99.24\%     & 99.24\%  & 98.47\%    & 98.47\%    & 98.6\%      & 98.47\%  & 98.65\%    & 98.65\%    & 98.66\%     & 98.65\%  \\
1730      & 98.72\%    & 98.71\%    & 98.74\%     & 98.72\%  & 98.61\%    & 98.61\%    & 98.63\%     & 98.61\%  & 97.57\%    & 97.57\%    & 97.65\%     & 97.57\%  \\ \hline
\end{tabular}}
\end{adjustbox}
\end{table}
\unskip

In table \ref{tab4}, different N2FGAN architectures are compared, the first condition is 1797 RPM, and the second condition is 1772 RPM. All experiments are done by considering 40000 steps for training the network. Three different architectures are considered for the generator (G) with 3,4, and 5 blocks and the discriminator(D) with 2,3, and 4 blocks. The number of neurons is mentioned in the table. These blocks are connected to make the networks. The results show that the deep networks perform better, and the training time is longer. Different input data lengths of 256,512  and 1024 samples are also studied. According to the table, N2FGAN can generate more accurate data while the input data size is large enough, and it performs better with the length of 512 and 1024. No significant difference in performance was found between these two last lengths.

% \begin{landscape}
\begin{landscape}
\begin{table}
\caption{Comparison between different architectures of the N2FGAN tested for 1772RPM.\label{tab4}}
\begin{adjustbox}{width=\columnwidth,center}
\newcolumntype{C}{>{\centering\arraybackslash}X}
\resizebox{0.9\paperwidth}{!}{\begin{tabular}{llllllllllllllll}
              &                                &                      &            & \multicolumn{4}{l}{ConvLSTM Classifer}   & \multicolumn{4}{l}{CNN Classifer}        & \multicolumn{4}{l}{ConvAE Classifer}     \\ \hline
Runtime(s) & Generator Blocks               & Discriminator Blocks & Input size & Accuracy & $F_1$ score & Precision & Recall & Accuracy & $F_1$ score & Precision & Recall & Accuracy & $F_1$ score & Precision & Recall \\ \hline
535.71   & 3(Input length-256-64)         & 2(64-256)            & 256        & 90.94\%    & 90.37\%    & 91.81\%     & 90.94\%  & 92.57\%    & 92.45\%    & 92.92\%     & 92.6\%   & 92.50\%     & 92.48\%    & 92.79\%     & 92.50\%   \\
682.18   & 3(Input length-256-64)         & 2(64-256)            & 512        & 98.54\%    & 98.53\%    & 98.60\%      & 98.40\%   & 97.67\%    & 97.66\%    & 97.75\%     & 97.67\%  & 91.87\%    & 91.57\%    & 93.79\%     & 91.87\%  \\
1282.18   & 3(Input length-256-64)         & 2(64-256)            & 1024       & 99.2\%     & 99.20\%     & 99.23\%     & 99.20\%   & 99.72\%    & 99.72\%    & 99.73\%     & 99.72\%  & 99.34\%    & 99.34\%    & 99.36\%     & 99.34\%  \\
674.56   & 4(Input length-256-128-64)     & 3(64-128-256)        & 256        & 94.44\%    & 94.35\%    & 94.81\%     & 94.44\%  & 92.20\%     & 92.01\%    & 93.00\%        & 92.19\%  & 88.89\%    & 88.74\%    & 90.16\%     & 88.89\%  \\
1346.31   & 4(Input length-256-128-64)     & 3(64-128-256)        & 512        & 98.78\%    & 98.78\%    & 98.81\%     & 98.78\%  & 98.85\%    & 98.85\%    & 98.89\%     & 98.85\%  & 99.27\%    & 99.70\%     & 99.28\%     & 99.27\%  \\
1381.87   & 4(Input length-256-128-64)     & 3(64-128-256)        & 1024       & 98.10\%     & 98.05\%    & 98.21\%     & 98.06\%  & 99.83\%    & 99.83\%    & 99.83\%     & 99.83\%  & 86.60\%     & 84.12\%    & 92.16\%     & 86.60\%   \\
775.10   & 5(Input length-512-256-128-64) & 4(64-128-256-512)    & 256        & 81.11\%    & 74.74\%    & 71.00\%        & 81.11\%  & 81.11\%    & 74.96\%    & 71.36\%     & 81.11\%  & 78.37\%    & 72.21\%    & 69.11\%     & 78.37\%  \\
1102.08   & 5(Input length-512-256-128-64) & 4(64-128-256-512)    & 512        & 99.24\%    & 99.23\%    & 99.25\%     & 99.24\%  & 98.26\%    & 98.26\%    & 98.35\%     & 98.26\%  & 96.15\%    & 96.11\%    & 96.74\%     & 96.15\%  \\
1812.25   & 5(Input length-512-256-128-64) & 4(64-128-256-512)    & 1024       & 99.72\%    & 99.72\%    & 99.72\%     & 99.72\%  & 98.04\%    & 98.38\%    & 98.48\%     & 98.4\%   & 88.02\%    & 86.10\%     & 92.20\%      & 88.02\%  \\ \hline
\end{tabular}}
\end{adjustbox}
\end{table}
\unskip
\end{landscape}

 In order to evaluate the performance of N2FGAN compared to other similar algorithms, a comparison panel including classical augmentation, and two state-of-the-art generative algorithms, Wasserstein GAN with gradient pentalty (WGAN-GP) and CGAN, are chosen. Classical augmentation is a set of operations such as reversing the signal burst, adding Gaussian noise to it, and flipping it by multiplying its values to minus one \cite{m16}. In this experiment, ConvLSTM is trained on all the six classes: As for the health class, it is trained on both motor speeds of 1797 and 1772 RPM. There are 250 training samples for the inner class, including 150 real samples for motor speed of 1797 and 100 samples generated using different augmentation frameworks in the comparison panel. As discussed earlier, N2FGAN takes the first condition (1797 RPM) and generates the second condition (1772 RPM), while the rest of the frameworks can only augment the first condition. As for the other four classes, only 150 real samples in the first condition are used to train the classifier. Table~\ref{tab5} exhibits the number of real and synthetic samples as well as the training condition and the test condition for different CWRU classes. In this experiment, the number of training samples for all the fault classes is set relatively low to resemble real-world situations where the practitioners have to deal with imbalanced and insufficient data.

\begin{figure}
\centering
\includegraphics[width=\textwidth]{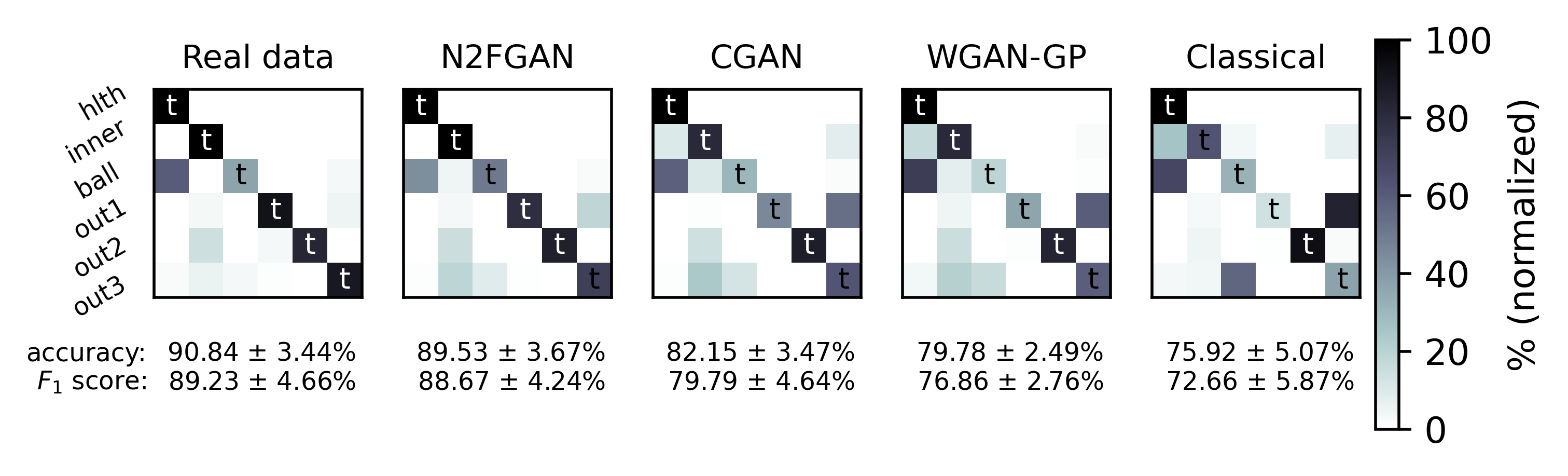}
% \end{adjustwidth}
\centering
\caption{Comparing the effect of each augmentation framework on the classifier performance when inner class is augmented - $t$ stands for true labeled samples. \label{fig10}}
\end{figure}

The experiment is run 20 times, each time with different training and test sets collected from the CWRU dataset. As it can clearly be seen in Figure~\ref{fig10}, N2FGAN outperforms the other augmentation frameworks. In fact, the results demonstrate that the performance of the classifiers trained on the real data and the data generated by N2FGAN were very similar. Classical augmentation had rather poor effectiveness since the average accuracy of the classifier trained on a non-augmented dataset is about 73.2\%, only 2\% less than classical augmentation. CGAN and WGAN-GP had mediocre effectiveness as they had only improved the accuracy by almost 9\% and 6\%, respectively.
\begin{table}
\caption{Training and test set configuration for comparing N2FGAN, CGAN, WGAN and classical augmentation.\label{tab5}}
\begin{tabular}{llllll}
\toprule
\multirow{2}{*}{classes} & \multicolumn{3}{c}{training set}                         & \multicolumn{2}{c}{test set} \\ \cline{2-4}
                         & RPM           & \#real samples & \#synthetic samples & RPM    & \#real samples  \\
\midrule
health                   & 1797 and 1772 & 3000           & 0                   & 1772   & 150             \\
inner                    & 1797          & 150            & 100                 & 1772   & 150             \\
ball                     & 1797          & 150            & 0                   & 1772   & 150             \\
outer1                   & 1797          & 150            & 0                   & 1772   & 150             \\
outer2                   & 1797          & 150            & 0                   & 1772   & 150             \\
outer3                   & 1797          & 150            & 0                   & 1772   & 150            \\
\bottomrule
\end{tabular}
\end{table}

% \clearpage

\section{Conclusions}
In industrial environments, fault data is scarce, and in many cases, normal data is abundant. Machines work in different conditions (i.e., numerous motor loads and speeds), for which fault samples are rarely available. This makes the utility of any machine learning based method limited since the developed model will be greatly biased to normal conditions. By augmenting normal data with sufficient fault data in a certain condition, the proposed framework enables machine learning based models that are more robust for fault diagnosis even in unforeseen fault conditions.

This paper introduces a novel data augmentation algorithm to synthesize fault data. In this algorithm, a variation of CGAN is proposed that can be trained on normal and fault data of one condition. Then the trained generator of the network is used to generate fault data from the normal samples for each motor speed for which there is no fault data available. The generated data is compared with the actual data and the normal input data using t-SNE. The results illustrate that the generated fault data has the same characteristics as the real fault data.

Moreover, three different classifiers are employed to validate the quality of the synthesized data. The classifiers are trained on various normal and actual fault samples. For the test phase, a new dataset is extracted from the primary dataset with the actual faults from the target class replaced by generated faults and fed into the trained classifiers for testing. In our experiments, three different conditions were tested with respect to different motor speeds. The results demonstrate that the generated faults are correctly classified with high accuracy (more than 97\% in all cases). This proves the generated fault data is very similar to the actual fault data.  On the other hand, three frameworks are provided (including CGAN and WGAN) to evaluate the effectiveness of the proposed model in an imbalanced condition. Compared to the others, N2FGAN has demonstrated a higher similarity to the real data and improved the classification performance significantly. 

Future extensions of the present work will focus on exploring the effectiveness of generating the signal features instead of the raw vibration samples. In addition, the work should explore an efficient hyperparameter tuning framework to train the generator faster without compromising its performance. Furthermore, reducing the complexity of the network to reduce the training time can be another venue for future work.

\bibliography{references}

\end{document}